\documentclass{article}
\usepackage{ifthen}
\usepackage{mdwlist}
\usepackage{amsmath,amssymb,amsfonts,amsthm}
\usepackage{bm}
\usepackage{bbm}

\usepackage[colorlinks=true,linkcolor=blue,citecolor=blue,urlcolor=blue]{hyperref}
\usepackage{graphicx}
\usepackage{xspace}
\usepackage{verbatim}
\usepackage{algorithm}
\usepackage[margin=1in]{geometry}
\usepackage{color}
\usepackage{thm-restate}
\usepackage{latexsym}
\usepackage{epsfig}
\usepackage{wrapfig}
\usepackage[T1]{fontenc}

\usepackage{hyperref}       
\usepackage{watml}
\usepackage{natbib}
\usepackage{algorithmic}
\newcommand{\INPUT}{\item[\textbf{Input:}]}
\newcommand{\OUTPUT}{\item[\textbf{Output:}]}

\title{Demystifying Foreground-Background \\ Memorization in Diffusion Models\footnote{Authors JZD and YL have equal contribution. Authors YY, GK, AD, FB are listed in reverse alphabetical order.}}

\author{
    Jimmy Z.\ Di\thanks{Cheriton School of Computer Science, University of Waterloo and Vector Institute. \texttt{jimmy.di@uwaterloo.ca}.}\hspace{0.8in}
    \and
    Yiwei Lu\thanks{University of Ottawa. \texttt{yiwei.lu@uottawa.ca}. }\hspace{0.8in}
    \and
    Yaoliang Yu\thanks{Cheriton School of Computer Science, University of Waterloo and Vector Institute. \texttt{yaoliang.yu@uwaterloo.ca}.}
    \and
    Gautam Kamath\thanks{Cheriton School of Computer Science, University of Waterloo and Vector Institute. \texttt{g@csail.mit.edu}. Supported by a Canada CIFAR AI Chair, an NSERC Discovery Grant, and an Ontario Early Researcher Award.}\hspace{0.25in}
    \and
    Adam Dziedzic\thanks{CISPA Helmholtz Center for Information Security. \texttt{adam.dziedzic@cispa.de}. Supported by the Deutsche Forschungsgemeinschaft (DFG, German Research Foundation), Project number 545047250.}\hspace{0.25in}
    \and
    Franziska Boenisch\thanks{CISPA Helmholtz Center for Information Security. \texttt{boenisch@cispa.de}.  Supported by the Deutsche Forschungsgemeinschaft (DFG, German Research Foundation), Project number 550224287.}
}

\begin{document}
\maketitle

\begin{abstract}

Diffusion models (DMs) memorize training images and can reproduce near-duplicates during generation. Current detection methods identify verbatim memorization but fail to capture two critical aspects: quantifying partial memorization occurring in small image regions, and memorization patterns beyond specific prompt-image pairs. To address these limitations, we propose Foreground Background  Memorization (\emph{FB-Mem}), a novel segmentation-based metric that classifies and quantifies memorized regions within generated images. Our method reveals that memorization is more pervasive than previously understood: (1) individual generations from single prompts may be linked to clusters of similar training images, revealing complex memorization patterns that extend beyond one-to-one correspondences; and (2) existing model-level mitigation methods, such as neuron deactivation and pruning, fail to eliminate local memorization, which persists particularly in foreground regions. Our work establishes an effective framework for measuring memorization in diffusion models, demonstrates the inadequacy of current mitigation approaches, and proposes a stronger mitigation method using a clustering approach.

\end{abstract}


\section{Introduction}

\begin{figure*}[h!]
\centering
\includegraphics[width=\linewidth]{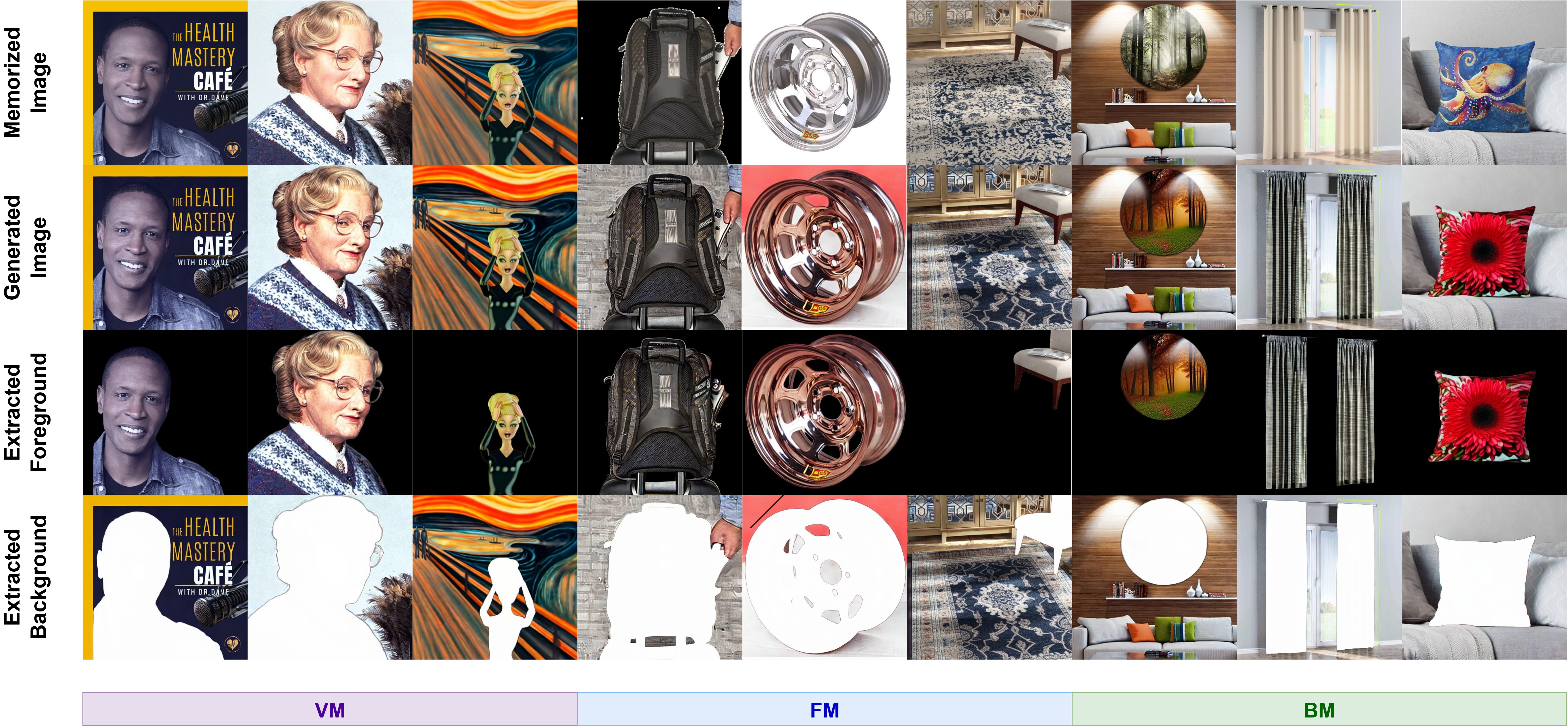}

\caption{\textbf{Examples of different types of memorization under our \emph{FB-Mem} evaluation.} We extract the foreground and background of the memorized images and generated images, and classify memorization using \Cref{alg:FB-Mem} to: Verbatim Memorization (column 1-3), Foreground Memorization (column 4-6), and Background Memorization (column 7-9).}
\label{fig:segmentation}
\vspace{-1.5em}
\end{figure*}

Diffusion models (DMs) \citep{SohlWMG15, song2020denoising, song2020score} and their text-to-image derivatives (e.g., Latent Diffusion \cite{RombachBLEO22}, DALL-E \citep{ramesh2022hierarchical}) have emerged as powerful generative frameworks, achieving remarkable success in producing high-fidelity images. Trained predominantly on large-scale datasets scraped from the Internet, such as LAION-5B \citep{SchuhmannBVGWCCKMW22}, these models often inherit both the richness and the risks associated with such data sources. A growing concern among researchers and practitioners is the potential for DMs to memorize and inadvertently reproduce portions of their training data—raising serious privacy, ethical, and legal issues \citep{CarliniHNJSTBIW23, SimepalliSGGG23a, SimepalliSGGG23b}. 
These issues become particularly problematic when reproduced content includes copyrighted materials or sensitive personal information without explicit consent or safeguards.

Existing detection methods (e.g., \citet{carlini2021extracting, SimepalliSGGG23a, wen2024detecting}) can accurately identify exact duplications using metrics like SSCD scores \citep{pizzi2022selfsuperviseddescriptorimagecopy} or CLIP scores \citep{radford2021learningtransferablevisualmodels}. While \citet{webster2023reproducibleextractiontrainingimages} and \citet{chen2025exploringlocalmemorizationdiffusion} have identified ``template memorization'' and ``local memorization'' respectively to detect partial memorization, it remains unclear how to quantify or measure the potential harm of such memorization. For instance, memorizing a color palette in the background of an image poses significantly less risk than memorizing a copyrighted object or identifiable feature.

In this work, we argue that inexact memorization detection should be more fine-grained to better assess the severity of partial memorization. Specifically, we propose  Foreground Background Memorization (\emph{FB-Mem}), a novel segmentation-based metric that classifies and quantifies memorized content across different regions of generated images. Given two images, \emph{FB-Mem} applies segmentation maps \citep{zheng2024bilateralreferencehighresolutiondichotomous} to differentiate foreground and background regions, compares each component using a pixel-wise image similarity metric 
and classifies the memorization into four categories: VM (verbatim memorization), FM (foreground memorization),  BM (background memorization), and NM (not memorized). 

Moreover, existing detection methods evaluate memorization for specific prompt-image pairs, neglecting the fact that DMs can generate diverse outputs from the same text prompt. Under \emph{FB-Mem} evaluation, we observe that these varied outputs are not necessarily linked to a single training image, but instead exhibit a one-prompt-to-many-training-images (\emph{one-to-many}) correspondence. We note that this notion is similar to ``retrieval verbatims'' by \citet{webster2023reproducibleextractiontrainingimages}. However, \citet{webster2023reproducibleextractiontrainingimages} only considers verbatim memorization and does not perform systematic analysis, a gap we aim to address in this work.  By clustering semantically similar text prompts, we can fully characterize this complex memorization behavior across different categories of memorization type using \emph{FB-Mem}.

Our proposed tools do not only establish an effective framework for detecting memorization in DMs, but also provide a robust evaluation methodology for assessing mitigation algorithms.
To address memorization, various mitigation strategies have been proposed, including inference-stage methods that adjust text-embedding or attention logits \citep{ren2024unveiling,wen2024detecting}, training-stage approaches that fine-tune pre-trained models \citep{ren2024unveiling,wen2024detecting}, and neuron-level interventions \citep{HintersdorfSKDB24,ChavhanBZLH24}. In this work, we focus on methods that permanently modify model weights \citep{wen2024detecting,HintersdorfSKDB24,ChavhanBZLH24}, which represent more fundamental and responsible changes to the model.
While such mitigation methods are effective against verbatim memorization, we observe that partial memorization, particularly foreground memorization, still persists after these mitigations. Furthermore, the \emph{one-to-many} correspondence also remains intact following these interventions.

In summary, we make the following contributions:
\begin{itemize}
\item We propose \emph{FB-Mem}, a segmentation-based metric that effectively detects and quantifies partial memorization in diffusion models beyond existing verbatim detection;
\item We reveal that memorization is more pervasive than previously understood, with individual generations linking to multiple training images and local memorization persisting after previous mitigation methods;
\item We demonstrate the inadequacy of current mitigation approaches using \emph{FB-Mem} and propose a novel clustering-based mitigation method.
\end{itemize}
\section{Background and Related Work}

\label{sec:background}

\paragraph{Neural network memorization:}
Neural network memorization is a common phenomenon in supervised learning \citep{arpit2017closer}, self-supervised learning \citep{wang2024LocalizeMemorizationSSL,wang2024memorization}, including contrastive learning~\citep{wang2025captured}, image auto-regressive models~\citep{kowalczuk2025privacyIARs}, and diffusion models \citep{SimepalliSGGG23a,SimepalliSGGG23b, wen2024detecting}. While it has been shown that memorization improves model generalization \citep{Feldman20, FeldmanZ20, wang2024memorization}, it could also lead to critical privacy concerns, such as data extraction attacks \citep{carlini2019secret, carlini2021extracting, carlini2023extracting,kowalczuk2025privacyIARs}. To detect such memorization in diffusion models, various approaches have been proposed, including SSCD scores \citep{pizzi2022selfsuperviseddescriptorimagecopy}, CLIP scores \citep{RombachBLEO22},  pairwise SSIM scores between initial noise differences \citep{webster2023reproducibleextractiontrainingimages}, distribution of attention \citep{ren2024unveiling}, edge inconsistency \citep{webster2023reproducibleextractiontrainingimages}, and predicted noise magnitudes \citep{wen2024detecting}.

We note that \citet{chen2025exploringlocalmemorizationdiffusion} apply brightened attention masks to identify local memorization, which shares similarities with our approach. However, we emphasize several key differences: (1) we provide a finer-grained classification that distinguishes between foreground and background memorization; (2) we conduct instance-level evaluation to identify one-prompt-to-many-training-images correspondence; and (3) we focus on model-level mitigation methods rather than prompt-level interventions.

\paragraph{Mitigating memorization:} To address the problem of memorization for DMs, various methods have been proposed. For example, inference-stage mitigation, including attention logit rescaling \citep{ren2024unveiling} and text-embedding adjustment \citep{wen2024detecting}; training-stage mitigation by fine-tuning an existing DM-based model \citep{ren2024unveiling,wen2024detecting}; and neuron-level mitigation \citep{MainiMSLKZ23, HintersdorfSKDB24,ChavhanBZLH24} that localizes and deactivates certain neurons responsible for memorization. In this work, we focus on mitigation methods that change the model parameters, including the fine-tuning approach \citep{wen2024detecting}, the neuron deactivation approach \citep{HintersdorfSKDB24}, and the weight pruning approach \citep{ChavhanBZLH24}.

Other techniques are also potentially applicable for mitigating memorization, for example, machine unlearning (proposed for removing private personal data) \citep{CaoYang15, BourtouleCCJTZLP21, SekhariAKS21, AldaghriMB21,WuLHH24} or concept removal (proposed for removing nudity or harmful concepts)~\citep{ChavhanLH24, GandikotaMKB23, LyuYHCJHXHD24} can potentially be used to remove the memorized information. 
\section{Measuring Memorization}
\label{sec:measure}

In this section, we (1) address the gap of partial memorization by proposing the novel Foreground Background Memorization (\emph{FB-Mem}) metric; and (2) propose an instance-level measurement for identifying \emph{one-to-many} correspondence.

\subsection{Memorization Pipeline} 
\label{sec:pipeline}
\paragraph{Existing approaches:}
Current research studies memorization in DMs, with particular focus on investigating memorization patterns in Stable Diffusion (SD) v1.4 \citep{RombachBLEO22}. While newer models exist, no datasets with memorized data are available for them. Consequently, state-of-the-art work \citep{webster2023reproducibleextractiontrainingimages,wen2024detecting,HintersdorfSKDB24} considers 500 memorized LAION prompts for SD v1.4, which we adopt in our paper.\footnote{We extend our discussion to Stable Diffusion 3 in \Cref{sec:exp_mit}.} \citet{HintersdorfSKDB24} further split this dataset into Verbatim Memorization (VM) and Template Memorization (TM) categories using SSCD scores \citep{pizzi2022selfsuperviseddescriptorimagecopy}, which represent the cosine similarity of image embeddings obtained from the Self-Supervised Copy Detection (SSCD) model. Specifically, \citet{HintersdorfSKDB24} uses a threshold of 0.7 to distinguish between these two classes.

\paragraph{A new pipeline:} Previous works typically assess memorization on a per-prompt basis, under the assumption that memorized prompts produce highly similar or even near-identical outputs across multiple generations. However, our empirical observations challenge this assumption: many memorized prompts result in significant variability across generated samples and may generate non-memorized samples even when using the same seed, as we will show in \Cref{sec:one-to-many}. Consequently, we manually reviewed and labeled 1,500 images generated using 300\footnote{We use this manually labeled subset to select the optimal similarity metric $M$ in \Cref{sec:fb-mem}, while we use the complete 500 prompts for subsequent experiments.} of the memorized prompts from \citet{chen2025exploringlocalmemorizationdiffusion}. Each image was reviewed and labeled as VM, TM, or NM based on its visual resemblance to ground-truth images. We show examples in \Cref{fig:manual_demo}, and the complete manual labeling protocol is described in detail in \Cref{app:label}.

\subsection{Measuring partial memorization}
\label{sec:fb-mem}

Previously, we discussed that existing approaches classify memorization into VM and TM.
While VM has a clear definition as exact duplication, TM lacks a rigorous definition. \citet{HintersdorfSKDB24} and  \citet{webster2023reproducibleextractiontrainingimages} state that TM reproduces only the general composition of training images while exhibiting non-semantic variations at fixed image positions. However, this definition does not accurately assess the potential harm of such memorization. For instance, memorizing a common background pattern poses minimal risk, whereas memorizing copyrighted content or artwork, even when appearing in a small region, should be identified and addressed (See \Cref{fig:segmentation} for examples). Motivated by this limitation, we aim to provide a fine-grained classification of TM that better captures the varying degrees of potential harm associated with different types of memorization.

\begin{algorithm}[t]
\caption{Foreground Background Memorization}
\label{alg:FB-Mem}
\begin{algorithmic}[1]
\INPUT Generated image $\xv_g$, training image $\xv_t$, similarity metric $M$, score threshold $\tau$, segmentation threshold $\beta$
\OUTPUT Memorization type $\in \{\text{VM, FM, BM, NM}\}$
\STATE Extract foreground mask $S_f(\xv_g),S_f(\xv_t)$; extract background mask $S_b(\xv_g),S_b(\xv_t)$
\STATE $M_{\text{full}} \leftarrow M(\xv_g, \xv_t)$

\IF{$\frac{|S_f|}{|x_g|} \leq \beta$}
\STATE $M_{\text{fg}} \leftarrow M(\xv_g, \xv_t \odot S_f(\xv_t))$; $M_{\text{bg}} \leftarrow M(\xv_g \odot S_b(\xv_g), \xv_t \odot S_b(\xv_t))$
\ELSIF{$\frac{|S_f|}{|x_g|} \geq 1 - \beta$}
\STATE $M_{\text{fg}} \leftarrow M(\xv_g \odot S_f(\xv_g), \xv_t \odot S_f(\xv_t))$; $M_{\text{bg}} \leftarrow M(\xv_g , \xv_t \odot S_b(\xv_t))$
\ELSE
\STATE $M_{\text{fg}} \leftarrow M(\xv_g \odot S_f(\xv_g), \xv_t \odot S_f(\xv_t))$; $M_{\text{bg}} \leftarrow M(\xv_g \odot S_b(\xv_g), \xv_t \odot S_b(\xv_t))$
\ENDIF
\IF{$M_{\text{full}} \geq \tau$}
    \RETURN \textsc{VM} (Verbatim Memorization)
\ELSIF{$M_{\text{fg}} \geq \tau$}
    \RETURN \textsc{FM} (Foreground Memorization)
\ELSIF{$M_{\text{bg}} \geq \tau$}
    \RETURN \textsc{BM} (Background Memorization)
\ELSE
    \RETURN \textsc{NM} (Not Memorized)
\ENDIF
\end{algorithmic}
\end{algorithm}

\paragraph{Foreground  background memorization:} We propose a novel metric for measuring memorization called  Foreground Background Memorization (\emph{FB-Mem}).
Our \emph{FB-Mem} algorithm (Algorithm~\ref{alg:FB-Mem}) utilizes a three-step comparison: (1) \emph{foreground/background extraction}:
given a pair of a generated image $\xv_g$ and a training image $\xv_t$,  \emph{FB-Mem} first applies segmentation to both images to extract foreground and background masks $S_f$ and $S_b$; (2) \emph{computing similarity}: given a a similarity metric $M$, we calculate the full image similarity $M_{\text{full}}$ between the generated and training images, and the foreground/background similarity between their extracted foreground/background, respectively; (3) \emph{memorization classification}: finally, given a threshold $\tau$, we classify the memorization into four possible types. 

Specifically, if the full image similarity exceeds the threshold $\tau$, \emph{FB-Mem} returns Verbatim Memorization~(VM). Otherwise, it checks foreground and background similarities in sequence, returning Foreground Memorization (FM) or Background Memorization (BM), respectively, if either exceeds the threshold. If none of the similarity scores meet the threshold, it classifies the pair as Not Memorized (NM).

\begin{figure}[t]
\vspace{-1em}
  \begin{center}
    \includegraphics[width=0.7\textwidth]{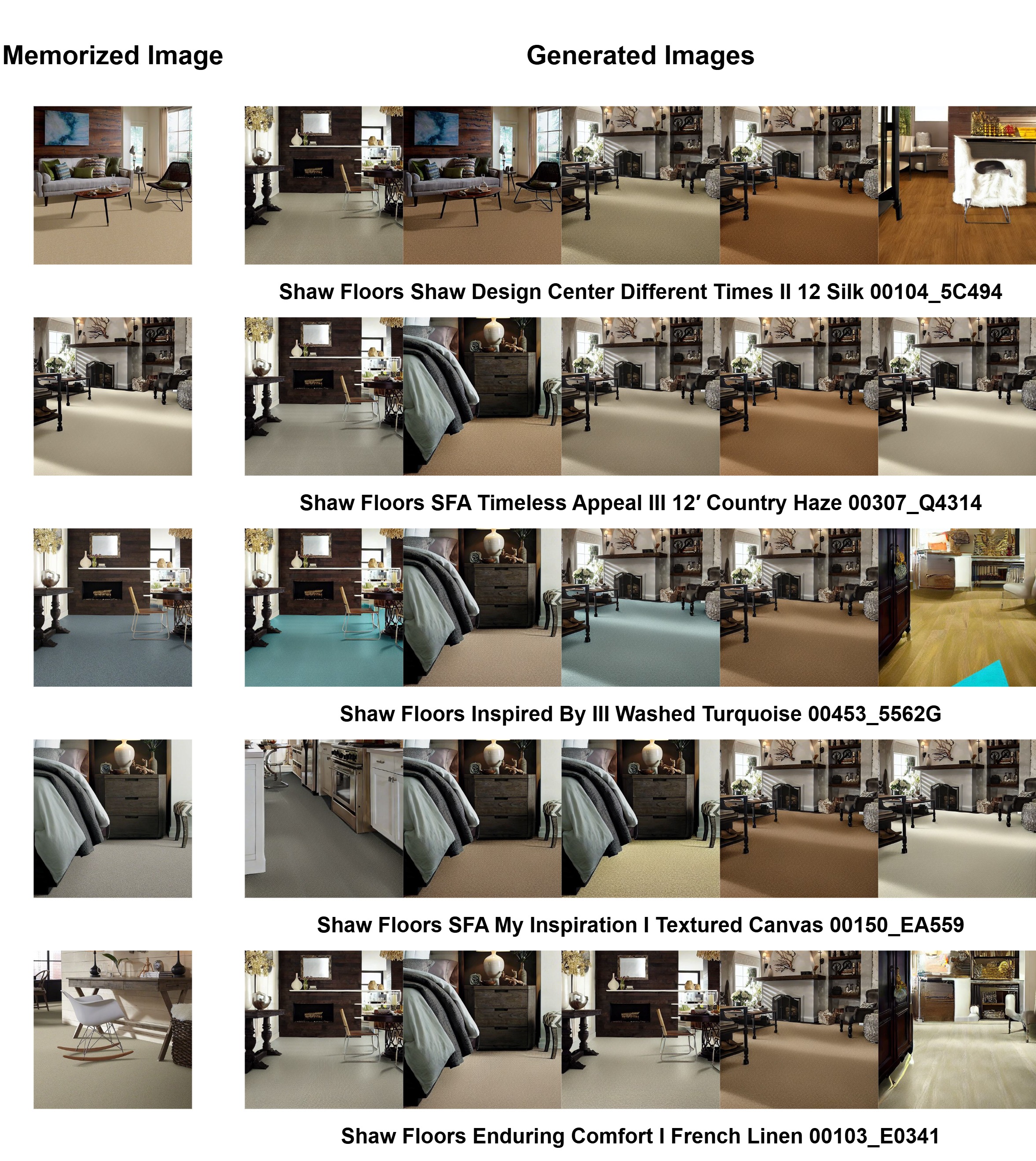}
    \vspace{-1em}
  \end{center}
    \caption{The memorized ground-truth image as well as the images generated using 5 different prompts selected from cluster 0 (i.e., the \textbf{Shaw Floors} cluster), in generation order.}
    \label{fig:prompts-clustering}
\end{figure}

To avoid reporting false positives when image segmentation fails, such as when the quality of the image is low, we perform an adaptive similarity computation based on the foreground proportion in the generated image. When the foreground region is very small (proportion $\leq \beta$ of the total image size, where $\beta$ is a tunable hyper-parameter), the algorithm compares the entire generated image against only the foreground of the training image for foreground similarity, while computing background similarity using masked regions from both images. Conversely, when the foreground dominates the image (proportion $\geq 1-\beta$), it compares the masked foreground regions but uses the entire generated image against the training image's background for background similarity. For balanced cases where the foreground proportion falls between these extremes, the algorithm performs standard masked comparisons for both foreground and background regions. For all experiments reported in this paper, we adapt the threshold of $\beta = 0.03$.

\paragraph{Choosing an optimal $M$:} In principle, our \emph{FB-Mem} algorithm can be equipped with any suitable similarity metric $M$. In this paper, we choose Multiscale Structural Similarity Index (MS-SSIM) \citep{1292216} as our metric $M$. SSIM is a standard tool for comparing pixel-wise image similarity through the lens of luminance, contrast, and structure. MS-SSIM further performs multiple re-scaling and down-sampling procedures on the contrast and structural components to obtain a more robust form. Details of SSIM and MS-SSIM are provided in \Cref{app:ssim}. Next we justify our choice of $M$.

\begin{table*}[t]
  \caption{Performance of each metric in classifying different memorization: 1) VM-NM; 2) TM-NM; 3) VM-TM.}
  \label{tab:metric_three_way_classification}
  \centering
  \scalebox{0.97}{\begin{tabular}{lllllllllll}
    \toprule
    \multirow{1}{*}{Classification} & \multicolumn{3}{c}{VM-NM} & \multicolumn{3}{c}{TM-NM} & \multicolumn{3}{c}{VM-TM} \\
    \cmidrule(lr){2-4} \cmidrule(lr){5-7} \cmidrule(lr){8-10}
    Metrics & SSIM & MS-SSIM & SSCD & SSIM & MS-SSIM & SSCD & SSIM & MS-SSIM & SSCD \\
    \midrule
    AUROC & 0.989 & 0.994 & \textbf{1.000} & 0.886 & 0.962 & \textbf{0.992} & 0.922 & \textbf{0.884} & 0.875 \\
    f-1 Score & 0.820 & \textbf{0.992} & 0.986 & 0.727 & 0.318 & \textbf{0.343} & 0.361 & \textbf{0.846} & 0.742 \\
    TP@1\%FP & 0.897 & 0.986 & \textbf{0.997} & 0.300 & 0.856 & \textbf{0.993} & 0.517 & \textbf{0.528} & 0.489 \\
    Accuracy & 0.954 & \textbf{0.997} & 0.995 & 0.779 & 0.651 & \textbf{0.661} & 0.404 & \textbf{0.913} & 0.831 \\
    \bottomrule
  \end{tabular}}
\end{table*}

\begin{figure}
\vspace{-1em}
  \begin{center}    \includegraphics[width=0.9\textwidth]{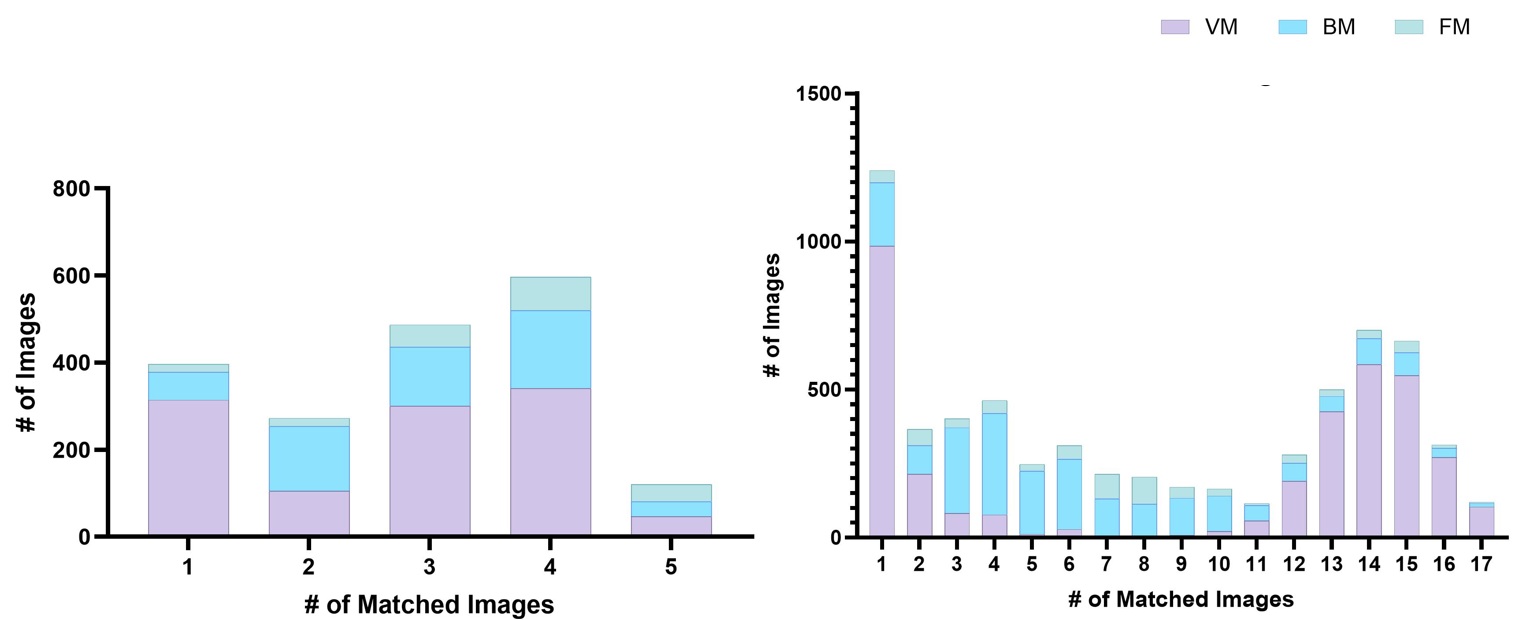}
    \vspace{-1em}
  \end{center}
    \caption{\textbf{Quantitative analysis of memorization in generated images} measured by \emph{FB-Mem}. The x-axis represents the number of distinct ground-truth images matched using each of the N images generated by each prompt, while the y-axis indicates the number of generated images identified as memorized by \emph{FB-Mem}. Each generated image produces exactly one match with one ground-truth image with the highest similarity score. Non-Memorized images are not included in the figure. \textbf{Left:} Default setting with $N=5$. \textbf{Right:} Ablation study with $N=20$.}
    \label{fig:ablation-n20}
    \vspace{-1em}
\end{figure}

\paragraph{Justifying the choice of $M$:} To evaluate the effectiveness of different memorization metrics, we conducted a three-way classification task (VM, TM,\footnote{In particular, we perform direct comparison of entire images (as done by previous work) rather than considering any sort of segmentation (as we prescribe in \Cref{alg:FB-Mem}).} NM) using our manually labeled dataset described in \Cref{sec:pipeline}.

We generated 1,500 images using the 300 labeled prompts and computed the similarity of each generated image to all 498\footnote{Two images were unavailable due to broken URLs.} ground-truth memorized images using three metrics: SSIM, MS-SSIM, and SSCD. For each generated image, we identified the highest-scoring ground-truth pair across all comparisons under each metric.

Similarity classification thresholds were established based on prior work. For SSIM and MS-SSIM, we set the VM threshold at 0.8 and the TM threshold at 0.6. For SSCD, we adopted settings from \citet{HintersdorfSKDB24} and \citet{wen2024detecting}, using a verbatim memorization (VM) threshold of 0.7 and a template memorization (TM) threshold of 0.5. Classification performance is reported in Table~\ref{tab:metric_three_way_classification}.

Our results demonstrate that while SSCD effectively classifies verbatim and non-memorized samples, it exhibits vulnerability to localized dissimilarities. Specifically, SSCD may fail to detect memorization when images contain small differing regions despite being visually near-identical overall.\footnote{This aligns with recent findings by \citet{chen2025exploringlocalmemorizationdiffusion}, which highlight SSCD's vulnerability to local perturbations.}
We also evaluated the accuracy-efficiency tradeoff across metrics. On an NVIDIA A6000 GPU, comparing a single generated image against 500 ground-truth training images requires over 5 minutes using SSCD, compared to only 24 seconds using MS-SSIM. This substantial computational advantage, combined with MS-SSIM's robustness to minor local variations, motivated our choice to build FB-Mem upon MS-SSIM.


\subsection{Measuring \emph{one-to-many} correspondence}
\label{sec:one-to-many}

In \Cref{sec:pipeline}, we note that existing methods neglect generation variations and only consider prompt-wise memorization. Using our instance-wise pipeline and \emph{FB-Mem}, we observe an intriguing phenomenon of one-prompt-to-many-training-images (\emph{one-to-many}) correspondence. Specifically, we perform $N=5$ generations per prompt and group the results into 5 categories according to the number of matched training images within the memorized dataset. For each category, we count the occurrences of VM, FM, and BM and present the results in \Cref{fig:ablation-n20} (left). We observe that a substantial number of prompts exhibit \emph{one-to-many} correspondence, demonstrating a variety of memorization types across generations from the same prompt.

\paragraph{Prompts Clustering:}
\label{sec:prompt_cluster}

Moreover, we observe that \emph{one-to-many} correspondence is not random but exhibits semantic coherence around shared concepts. To quantify this behavior, we first encode each prompt into an embedding using the CLIP-ViT-B model, then apply K-Nearest Neighbors (KNN) clustering to group the 500 prompts into 12 distinct clusters. Figure~\ref{fig:prompts-clustering} shows examples of five prompts sampled from the same cluster, along with their corresponding generated images.\footnote{Notably, some generated images do not appear to be direct copies of training data. 
Interestingly, when we increase the number of $N$ in the next paragraph, we are able to generate replicates of these images as well.} Additional examples are provided in \Cref{app:add_exp}.

\paragraph{Ablation Study on $N$:}

Finally, we conduct an ablation study with an increased number of generated images per prompt ($N=20$) and present the results in \Cref{fig:ablation-n20} (right). We observe that prompts exhibiting \emph{one-to-one} correspondence remain roughly the same as with $N=5$. However, when we increase $N$, \emph{one-to-many} correspondence can extend up to 17 matching images, demonstrating the capability of diffusion models to memorize a large number of training images within a single prompt.

\section{Mitigating Memorization}
In the previous section, we established a new memorization pipeline, proposed \emph{FB-Mem} for memorization evaluation, and identified the phenomenon of \emph{one-to-many} correspondence. In this section, we (1) examine existing model-based mitigation methods under the \emph{FB-Mem} framework; (2) propose a cluster-wise mitigation approach; and (3) design a scoring metric for mitigation evaluation and analyze the utility-quality tradeoff in post-mitigation models.

\subsection{Clustering-based Mitigation}
\label{sec:mitigation}
\begin{figure}[t]
    \centering
    \includegraphics[width=0.9\linewidth]{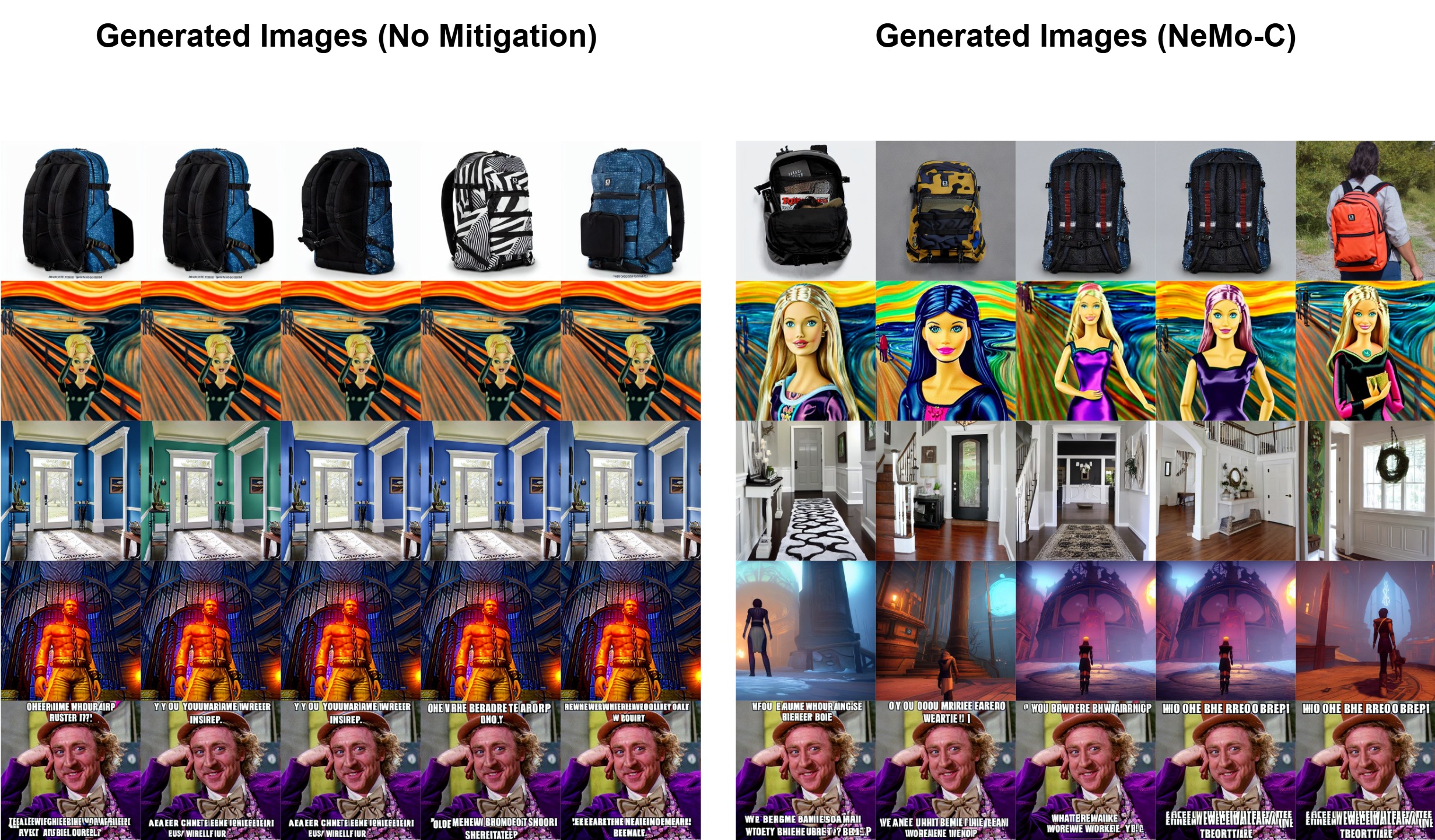}
    \caption{\textbf{Examples of generated images before and after applying NeMo-C.} The provided prompts, from top to bottom, are: (1) \emph{ALPHA Convoy 320 Backpack - View 4}; (2) \emph{If Barbie Were the Face of the World's Most Famous Paintings}; (3) \emph{Foyer Painted in WHITE}; (4) \emph{Dreamfall Chapters: The Longest Journey Will Be a PlayStation 4 Exclusive}; (5) \emph{Willy Wonka - Oh, you are in IB? Please tell me how much smarter you are than everyone else}. }
    \label{fig:nemoc}
\end{figure}

In the previous section, we observed the phenomenon of \emph{one-to-many} correspondence, which suggests that memorization occurs at the concept level rather than the prompt level. We therefore propose prompt clustering to quantify this notion of concept. This approach naturally extends to a stronger mitigation strategy, where we address clusters of memorization collectively.

Specifically, we build upon NeMo \citep{HintersdorfSKDB24}, a prompt-wise mitigation approach due to its superior mitigation performance on prompt-wise mitigation and preserving utility. NeMo utilizes a two-step process: (1) \emph{initial selection}: for each prompt in a cluster, NeMo identifies a broad set of candidate neurons that may be responsible for memorizing a specific training image; and (2) \emph{refinement}: NeMo filters the initial candidate set to obtain a smaller, refined set of neurons for each prompt.

Our clustering-based mitigation approach (denoted as NeMo-C) introduces a third step of \emph{aggregation}: given a memorized cluster, we compute the union of all refined neuron sets across the cluster, resulting in a consolidated set of neurons that consistently contribute to memorization. For each prompt in the cluster, we deactivate all neurons in this union set to mitigate memorization.

Unlike NeMo, which deactivates neurons specific to individual prompts, NeMo-C performs mitigation using the aggregated union set of neurons across the entire cluster. This enables robust memorization mitigation, specifically targeting the \emph{one-to-many} phenomenon described previously. In \Cref{sec:exp_mit}, we demonstrate that NeMo-C does not significantly decrease model utility compared to NeMo and other mitigation methods. We present selected examples of generated images before and after applying NeMo-C, including failure cases where generated images remain verbatim memorized post-mitigation, in Figure~\ref{fig:nemoc}. Additional examples are provided in Appendix~\ref{app:gen_examples}.

\subsection{Mitigation Evaluation}
Asides from \emph{FB-Mem}, we propose two metrics to evaluate memorization mitigation: the mitigation strength, quantified by a novel scoring function, and image quality post mitigation which measures model utility.

\begin{table}[h]
\caption{Mitigation Strength Scoring Function.}
\label{tab:scoring_function}
\centering
\scalebox{1}{\begin{tabular}{cc|cc|cc}
\toprule
\multicolumn{2}{c|}{\textbf{From VM}} & \multicolumn{2}{c|}{\textbf{From BM}} & \multicolumn{2}{c}{\textbf{From FM}} \\
\midrule
VM$\to$NM & +2.0 & BM$\to$FM & -0.5 & FM$\to$BM & +1.0 \\
VM$\to$BM & +1.5 & BM$\to$VM & -1.5 & FM$\to$NM & +1.5 \\
VM$\to$FM & +0.5 & BM$\to$NM & +0.5 & FM$\to$VM & -0.5 \\
\midrule
\multicolumn{6}{c}{\textbf{From NM}} \\
\midrule
\multicolumn{2}{c}{NM$\to$VM: -2.0} & \multicolumn{2}{c}{NM$\to$FM: -1.5} & \multicolumn{2}{c}{NM$\to$BM: -0.5} \\
\bottomrule
\end{tabular}}
\vspace{-1em}
\end{table}

\paragraph{Mitigation Strength:}

To evaluate the effectiveness of memorization mitigation methods, we introduce a scoring function in \Cref{tab:scoring_function} that quantifies the strength of mitigation based on memorization type transitions. Our scoring system assigns numerical values to transitions between different memorization states: Verbatim Memorization (VM), Background Memorization (BM), Foreground Memorization (FM), and Not Memorized (NM). The scoring function operates on the principle that transitions reducing memorization severity receive positive scores, while those increasing memorization or introducing new memorization patterns receive negative penalties. 

\paragraph{Image Quality:}
We use Q-Align \citep{wu2024qalign} and DB-CNN \citep{Zhang_2020} to evaluate the quality of images after applying memorization mitigation. Due to their no-reference (NR) nature, both methods perform well even with our relatively small sample size. Q-Align leverages aligned Vision-Language models (VLM) to assess generation quality, while the DBCNN method is based on Deep Bilinear Convolutional Neural Network. Both methods were implemented using the Image Quality Assessment (IQA) toolbox.

\subsection{Experimental Settings}

\paragraph{Baseline Methods:} We consider three model-based mitigation methods: NeMo \citep{HintersdorfSKDB24}, Wanda \citep{ChavhanBZLH24},\footnote{Note that Wanda was originally proposed by \citet{SunLBK23} for large language models, while we utilize its adaptation for diffusion models as presented in \citet{ChavhanBZLH24}.} and DetectMem \citep{wen2024detecting}.
All mitigation methods are applied using their default hyperparameters as specified in the respective papers. For NeMo, we use an activation threshold of 0.428 to determine which neurons to deactivate for each of the 500 prompts. This threshold corresponds to the mean plus one standard deviation of pairwise SSIM scores between initial noise differences, measured on a holdout set of 50,000 LAION prompts. The number of deactivated neurons varies across prompts, ranging from 0 to 436.
For Wanda, the sparsity threshold (i.e., the percentage of pruned neurons) is set to 1\%. For DetectMem, we evaluate both training-time and inference-time mitigation approaches. The training-time mitigation uses the pre-fine-tuned SD v1.4 model provided by the authors. For inference-time mitigation, we use the default configuration with a target loss of 3.

\paragraph{Evaluation Pipeline:} We use the 500 memorized prompts from \citet{webster2023reproducibleextractiontrainingimages} as our pre-mitigation benchmark. Using Stable Diffusion v1.4, we generate five images per prompt using one random seed without applying any mitigation techniques, following the same settings described in Section~\ref{sec:measure}. We then apply various mitigation methods, including baseline methods and our NeMo-C method,  and regenerate five images per prompt under each method. Each generated image is compared to all 498 of the retrievable memorized images, resulting in a total of 1,245,000 comparisons. Additionally, we include \textbf{(1)} \emph{FB-Mem} results for 500 randomly selected LAION prompts from the non-memorized set, generating 5 images per prompt as well as \textbf{(2)} 2500 images generated using the state-of-the-art Stable Diffusion 3 model using the 500 memorized prompts.

\begin{figure}
\vspace{-1em}
  \begin{center}
    \includegraphics[width=0.7\textwidth]{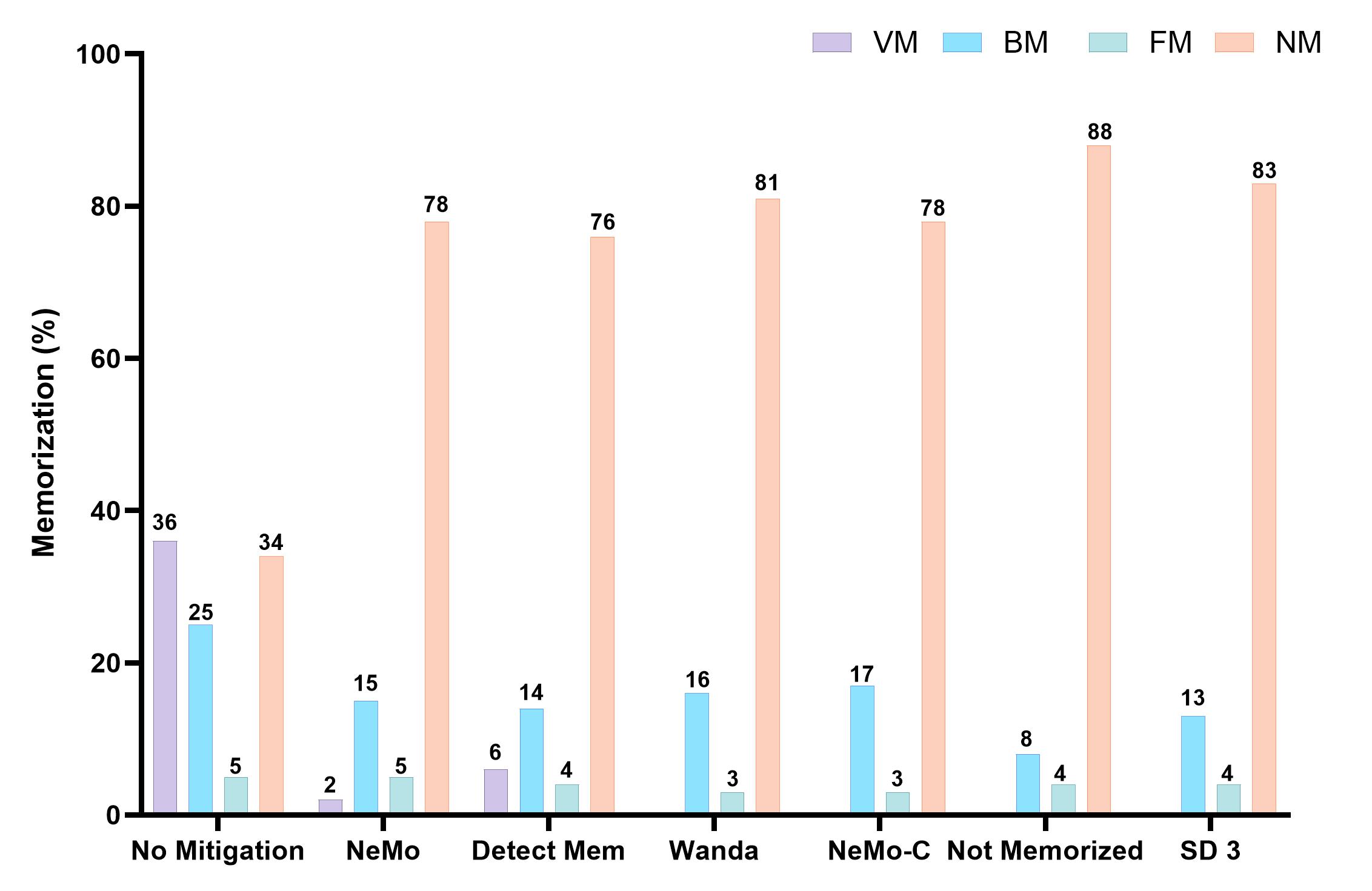}
    \vspace{-1em}
  \end{center}
  \caption{Memorization distribution evaluated using \emph{FB-Mem} before/after mitigation, not memorized prompts, and Stable Diffusion 3. }\label{fig:similarity_score_mitigation}
\end{figure}

\begin{table}[h] 
  \caption{Model performance assessed by image quality metrics before/after mitigation. The reported scores are computed for all 2,500 generated images and averaged.}
  \label{tab:image_quality_mitigation}
  \centering
  \begin{tabular}{lll}
    \toprule
    \multirow{1}{*}{Mitigation Method }  &
      \multicolumn{1}{c}{DB-CNN} &
      \multicolumn{1}{c}{Q-Align}
      \\
    \midrule
    Pre-mitigation  & 0.60 & 4.02\\
    NeMo & 0.587 & 3.63\\
    Detect Mem  & 0.449 &  3.41 \\
    Wanda&  0.578 & 3.55 \\ 
    NeMo-C  & 0.586 & 3.52\\
    \bottomrule
  \end{tabular}
  \vspace{-1em}
\end{table}
\subsection{Experimental Results}
\label{sec:exp_mit}

\begin{figure}[t]
    \includegraphics[width=\linewidth]{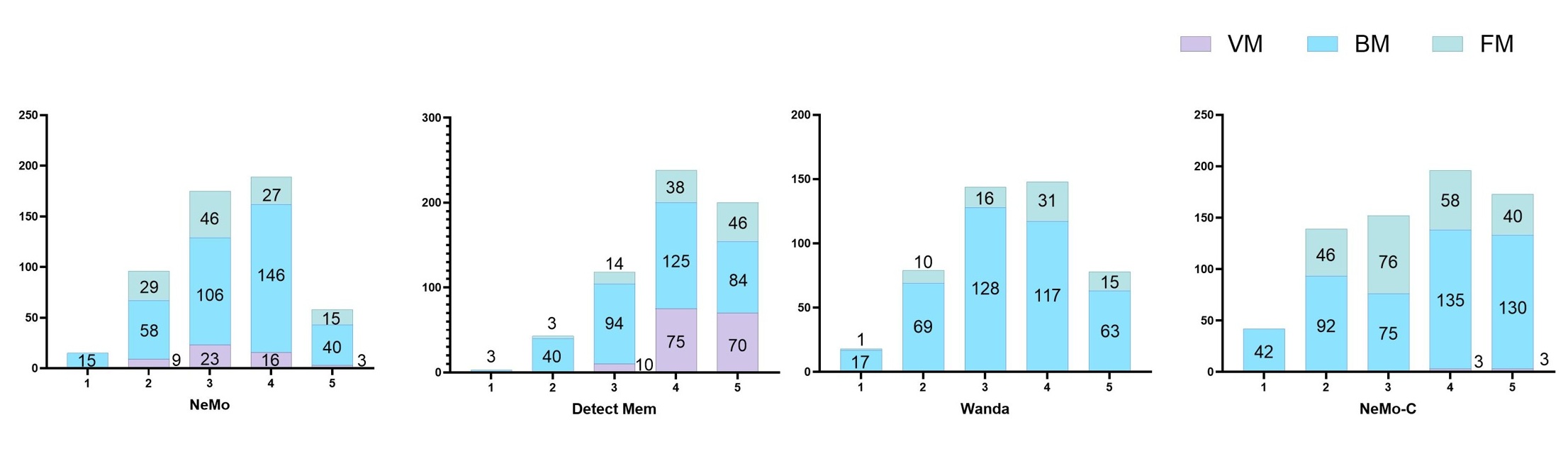}
\caption{\emph{One-to-many} correspondence after applying mitigation methods, measured using \emph{FB-Mem}.}
    \label{fig:4-dist}
\end{figure}

\paragraph{Memorization distribution under \emph{FB-Mem}:}
We first evaluate the effectiveness of each mitigation method using the \emph{FB-Mem} metric. As shown in \Cref{fig:similarity_score_mitigation}, most methods are highly effective at reducing verbatim memorization (VM), eliminating over 90\% of such cases. However, while background memorization (BM) is largely alleviated, foreground memorization (FM) persists.
Notably, we examine the memorized prompts on Stable Diffusion 3 and observe that memorization does not exhibit significantly even without mitigation,\footnote{We acknowledge that this might be because the memorized prompts differ significantly across different versions of Stable Diffusion, a problem we aim to study systematically in future work.} matching the observations of \citet{HintersdorfSKDB24}. Moreover, in \Cref{fig:4-dist}, we demonstrate that \emph{one-to-many} correspondence still largely persists across methods, with NeMo-C showing slight improvements over baselines.

\paragraph{Mitigation strength:} Although the memorization distribution analysis provides insights into post-mitigation results, it fails to capture the memorization transitions induced by different mitigation methods. To address this, we calculate the average mitigation score (across 2,500 images, higher scores indicate better performance) for each method according to the scoring function in \Cref{tab:scoring_function}. The results are: NeMo (0.74), DetectMem (0.67), Wanda (0.79), and NeMo-C (0.83). We observe that NeMo-C achieves the highest mitigation strength, demonstrating the effectiveness of our concept-wise clustering approach.

\begin{figure}[t]
\vspace{-1em}
  \begin{center}
    \includegraphics[width=0.55\textwidth]{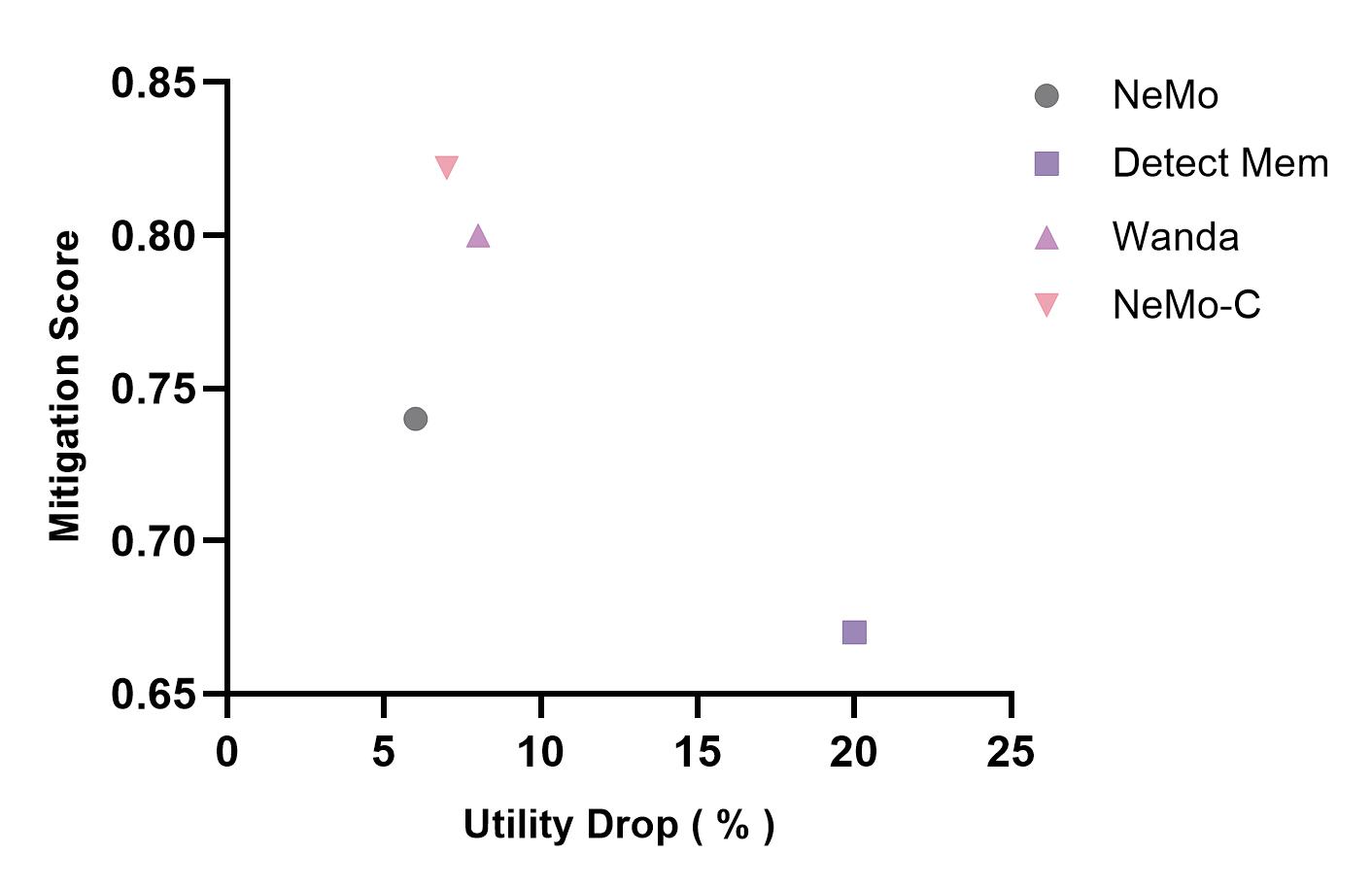}
    \vspace{-1em}
  \end{center}
\caption{Trade-off between utility and mitigation efficacy. The y-axis shows the mitigation score (higher is better), while the x-axis indicates the drop in image quality compared to the non-mitigated baseline (lower is better). The utility drop is calculated using the two methods shown in Table~\ref{tab:image_quality_mitigation}.}
    \label{fig:tradeoff}
\end{figure}

\paragraph{Utility trade-off:}

Finally, a key evaluation criterion for mitigation methods is performance preservation. In Table~\ref{tab:image_quality_mitigation}, we report the average quality of generated images after applying mitigation methods. Among all approaches, NeMo and NeMo-C achieve the best image quality. Moreover, we present the trade-off between mitigation strength and utility degradation in \Cref{fig:tradeoff}, where NeMo-C emerges as the optimal mitigation method.

Additionally, we demonstrate that the trade-off between mitigation efficacy and image quality can be further fine-tuned. For both NeMo and our proposed NeMo-C method (as described in this section), the identified set of neurons in the U-Net is completely deactivated. As an alternative, these neurons can be dampened rather than entirely disabled. Specifically, we apply a multiplicative dampening factor, $\alpha_{\text{damp}}$, to the selected neurons, where $\alpha_{\text{damp}} = 0$ corresponds to standard NeMo-C. As expected, increasing the dampening factor leads to improved image quality while slightly reducing mitigation efficacy. The extended NeMo-C method and detailed experimental settings are described in \Cref{sec:dampen}.


\section{Conclusion}

In this work, we addressed critical limitations in memorization detection for DMs by proposing \emph{FB-Mem}, a segmentation-based metric that provides fine-grained classification of memorized content. Our analysis revealed that memorization is fundamentally cluster-wise rather than prompt-wise, with individual generations incorporating content from multiple training images simultaneously.
Using the \emph{FB-Mem} framework, we demonstrated that existing mitigation methods fail to eliminate local memorization, particularly in foreground regions. Our proposed NeMo-C cluster-wise mitigation approach achieves more robust memorization reduction while maintaining model utility.

Our work establishes a proper measurement of the memorization pipeline of DMs and opens new directions, such as extending these findings to other generative modalities like large language models and developing more sophisticated semantic-based clustering and mitigation strategies regarding foreground memorization.



\newpage
\bibliographystyle{plainnat}
\bibliography{arxiv/output}  
\newpage
\appendix
\section{SSIM and Multiscale-SSIM}
\label{app:ssim}

First, we provide a brief overview of the Structural Similarity Index (SSIM) \citep{WangBSS04} and the improved Multiscale Structural Similarity Index (MS-SSIM) \citep{1292216}. Let $\mv = \left\{\mv_i | i=1, 2,..., N \right\}$ and $\nv = \left\{ \nv_i | i = 1, 2, ..., N \right\}$ be two pixel groups extracted from the same spatial location from two images being compared, and $\mu_\mv$, $\sigma_\mv^2$ and $\sigma_{\mv\nv}$ be the mean of $\mv$, variation of $\mv$, and the covariance of $\mv$ and $\nv$, respectively. Then, the standard SSIM score is calculated by looking at the luminance $(l)$, contrast $(c)$, and structure $(s)$ as follows:

\begin{align}
l(\mv, \nv) &= \frac{2\mu_\mv \mu_\nv + c_1}{\mu_\mv^2 + \mu_\nv^2 + c_1}\\
c(\mv, \nv) &= \frac{2\sigma_\mv \sigma_\nv + c_2}{\sigma_\mv^2 + \sigma_\nv + c_2}\\
s(\mv, \nv) &=  \frac{\sigma_{\mv\nv} + c_3}{\sigma_\mv\sigma_\nv + c_3}
\end{align}

where $c_1, c_2$, and $c_3$ are small constants to stabilize the division with a weak denominator. With three components set to equal importance, we obtain the SSIM score:
\begin{align}
 \texttt{SSIM}(\mv, \nv) = \frac{(2\mu_{\mv}\mu_{\nv}+c_1)(2\sigma_{\mv\nv}+c_2)}{(\mu_{\mv}^2+\mu_{\nv}^2+c_1)(\sigma_{\mv}^2+\sigma_{\nv}^2+c_2)}, 
 \label{eq:ssim}
\end{align}

Furthermore, by performing multiple re-scaling and down-sampling procedures to the contrast component and structural component using a scaling factor $j$, we can obtain a more robust form of SSIM. Let 1 be the scale of the original images and Scale K be the maximum scale, the Multi-scale Similarity Index is obtained through:
\begin{align}
\texttt{SSIM}(\mv, \nv) &= \left[l_K(\mv, \nv) \right]^{\alpha_M} \dot \displaystyle \prod^K_{j=1}\left[ c_j(\mv, \nv)\right]^{\beta_j}\left[ s_j(\mv, \nv)\right]^{\gamma_j}.
\end{align}

In practice, the relative importance hyperparameter $\alpha, \beta$, and $\gamma$ are normalized and set equal at all values of $j$.

\section{Additional Experiments}
\label{app:add_exp}
\subsection{Ablation Study on $N$}
In this section, we provide additional details and example images from the ablation study where the number of generations per prompt $N$ is increased to 20.

\paragraph{Experiment Setting:} 
Following the pre-mitigation experiments in \Cref{sec:measure}, we use the same 500 memorized prompts from \citet{webster2023reproducibleextractiontrainingimages}. Using Stable Diffusion v1.4 with a fixed random seed, we generate $N=20$ images per prompt without applying any mitigation techniques.

\begin{figure}[h]
    \centering
    \includegraphics[width=\linewidth]{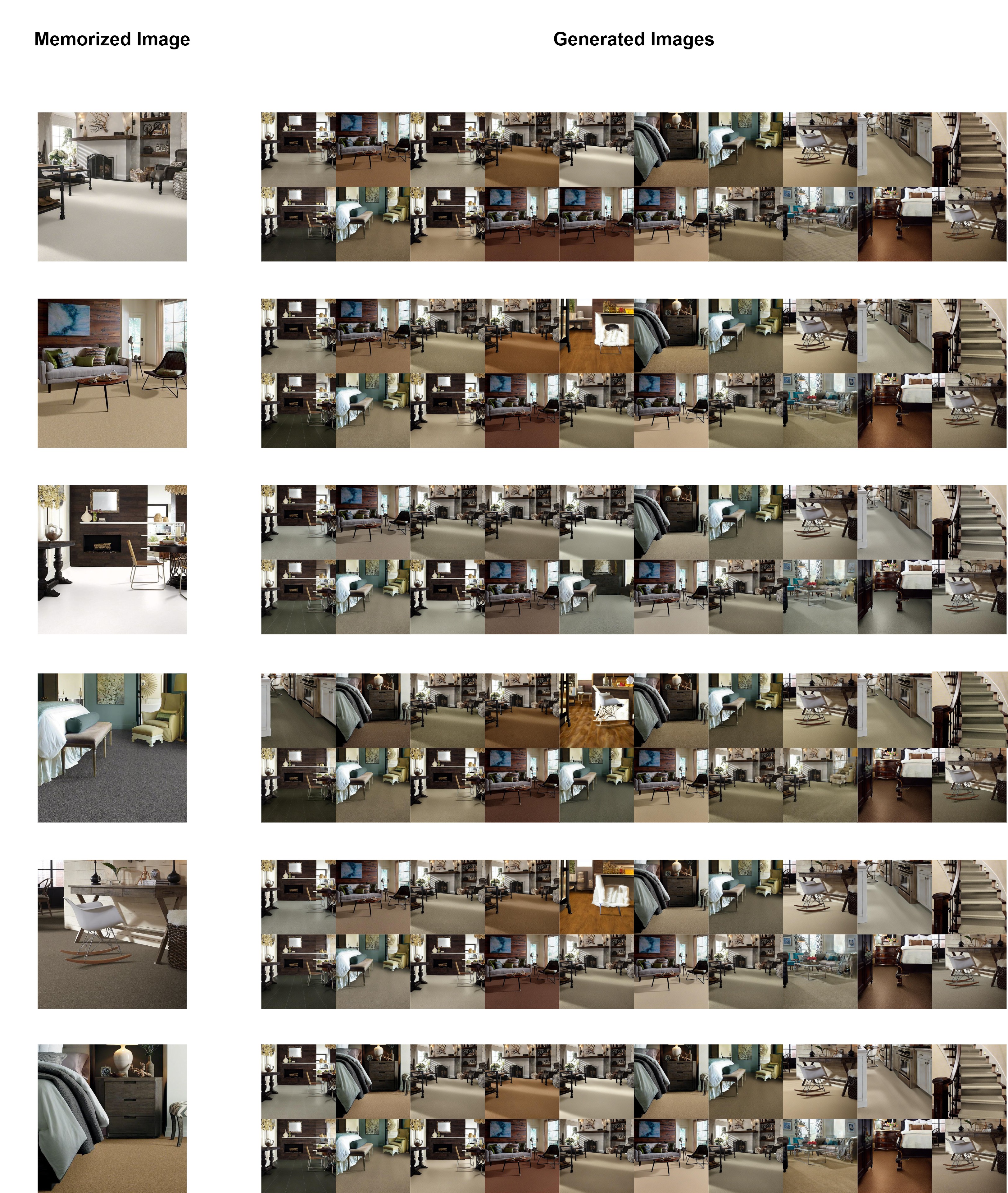}
    \caption{Twenty images generated across six different prompts selected from the \textbf{Shaw Floors} clusters.}
    \label{fig:N20-Extra}
\end{figure}

\paragraph{Clustering Example:} 
Figure~\ref{fig:N20-Extra} shows 20 images generated for six prompts from the \textbf{Shaw Floors} cluster. The first five prompts (rows 1–5) and their corresponding first five outputs were previously shown in \Cref{fig:prompts-clustering}. Notably, prompts 1–6 each produce nearly identical images beyond the initial five generations, maintaining the same sequential order. This observation provides strong evidence that the model clusters semantically similar training prompts and offers insights into how diffusion models internalize and reproduce content from their training data.

\subsection{NeMo-C with a Dampening Factor}
\label{sec:dampen}
For both NeMo \citep{HintersdorfSKDB24} and our proposed NeMo-C method (described in \Cref{sec:mitigation}), the refined set of neurons in the U-Net is completely deactivated. An alternative approach is to dampen these neurons rather than fully deactivating them. Specifically, we apply a multiplicative dampening factor $\alpha_{\text{damp}}$. In this section, we explore the effect of using such a dampening mechanism, experimenting with values $\alpha_{\text{damp}} = 0.1$ and $0.2$. We hypothesize that this approach may improve the visual quality of the generated images—by retaining some informative signal from the suppressed neurons—while partially sacrificing the mitigation strength compared to the original NeMo-C.

\begin{table}[h]
  \caption{Model performance assessed by mitigation efficacy and image quality metrics after applying mitigation with varying dampening factor $\alpha_{damp}$. When $\alpha_{damp} = 0$, we obtain the standard NeMO-C method described in the main paper.The reported mitigation score is calculated by aggregating 2500 scores using the method outlined in Table \ref{tab:scoring_function}. The reported quality scores are computed for all 2,500 generated images and averaged.}
  \label{tab:dampening}
  \centering
  \begin{tabular}{lll}
    \toprule
    \multirow{1}{*}{Dampening Factor ($\alpha_{damp}$) } &
      \multicolumn{1}{c}{Mitigation Score} &
      \multicolumn{1}{c}{Image Quality (DB-CNN)}
      \\
    \midrule
    0 (NeMo-C) & 0.83  & 0.586\\
    0.1 & 0.81 & 0.588 \\
    0.2 &  0.797 & 0.590 \\
    \bottomrule
  \end{tabular}
  \vspace{-1em}
\end{table}

The results presented in \Cref{tab:dampening} are consistent with our hypothesis. As the dampening factor increases, the image quality, measured by DB-CNN score, improves slightly, indicating that partial retention of neuron activations may help preserve useful generative capacity. However, this improvement in quality comes at the cost of a modest reduction in mitigation effectiveness. These findings suggest that dampening offers a tunable trade-off between mitigation strength and image fidelity, providing a flexible alternative to hard neuron deactivation.

\section{Generated Examples}
\label{app:gen_examples}

\begin{figure*}[h]
    \centering
    \includegraphics[width=0.9\linewidth]{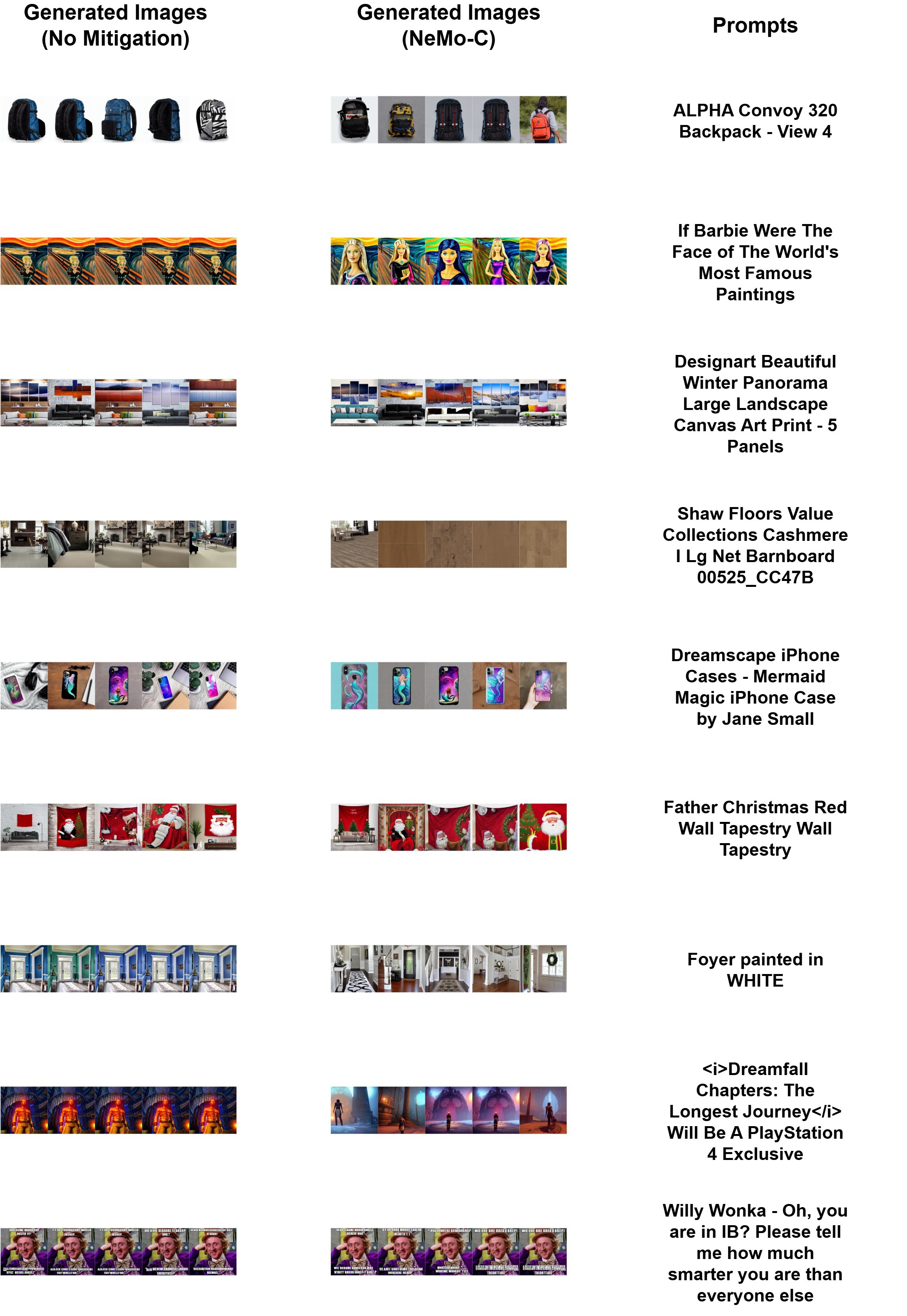}
    \caption{Distribution of generated more images before and after applying mitigation methods}
    \label{fig:nemoc-full}
\end{figure*}

In this section, we present additional examples of generated images along with their corresponding memorized training images and prompts, as shown in \Cref{fig:nemoc-full}. These examples include both successful mitigation cases and failure cases where NeMo-C fails to eliminate memorization.

The examples demonstrate that memorization can be harmful in several ways: \textbf{(1)} it reduces the diversity of generated outputs (rows 4–6); \textbf{(2)} it can lead to inaccurate generations that contradict prompt specifications (e.g., row 7, where the prompt explicitly requests a \textbf{white} foyer but the output displays blue or green tones); and \textbf{(3)} it raises potential copyright concerns (rows 2 and 8) as the generated images closely resemble artistic works by human creators used in games or other media.

However, memorization may sometimes benefit the model by generating highly accurate or detailed outputs that align well with user prompts—particularly when the underlying content is in the public domain, as illustrated in the final row. This dual nature of memorization highlights a promising direction for future research: developing methods to distinguish between harmful and benign memorization in diffusion models, similar to the approach taken by \citet{aerni2024measuringnonadversarialreproductiontraining} for language models.

\section{Manual Labeling}
\label{app:label}

As discussed in \Cref{sec:measure} of the main paper, existing methods are insufficient for accurately distinguishing between different types of memorization. This limitation motivates our proposed approach, \textit{FB-Mem}. To rigorously evaluate the effectiveness of this method, it is essential to establish reliable ground-truth labels that reflect human perception. Therefore, we manually labeled the images generated from memorized prompts (without applying mitigation techniques) and used these labels as the ground truth for our classification experiments in the main paper.

\begin{figure}[h]
\centering
\includegraphics[width=\linewidth]{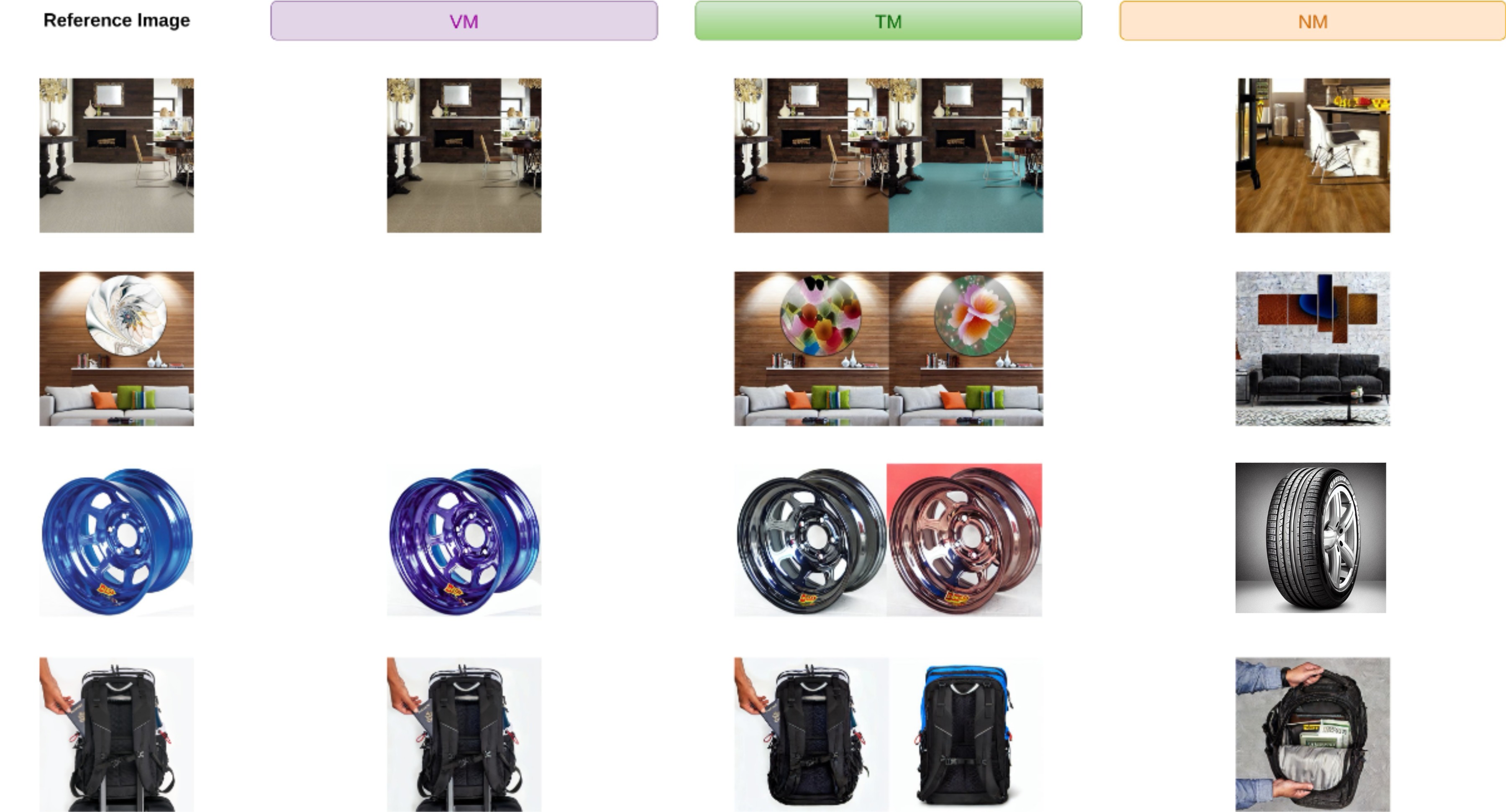}
\caption{Examples of manual classification for different reference images. For the reference image in row 2, no VM example was found.}
\label{fig:manual_demo}
\end{figure}

\paragraph{Labeling Procedure.}
We adopt a conservative approach when identifying verbatim memorization (VM)—an image is labeled as VM only if it is visually identical to its corresponding training image. For template memorization (TM), we consider both locally similar and dissimilar regions, as well as the overall visual style, following the examples illustrated in Figure \Cref{fig:manual_demo}. Each generated image is manually compared to up to ten of the most visually similar memorized images, as retrieved by BF-Mem and scored using SSCD, before reaching a labeling decision. Due to the time-intensive nature of this process, we labeled only the first 1,500 generated images, which we consider a sufficient sample for reliable evaluation.

\end{document}